# LLM-Generated Natural Language Meets Scaling Laws: New Explorations and Data Augmentation Methods

Zhenhua, Wang

School of Information Resource Management, Renmin University of China, zhenhua.wang@ruc.edu.cn

School of Computing, National University of Singapore, wangzhenhua@u.nus.edu

Guang Xu

School of Information Resource Management, Renmin University of China, 2020000919@ruc.edu.cn

Ming Ren*

School of Information Resource Management, Renmin University of China, renm@ruc.edu.cn

With the ascent of large language models (LLM), natural language processing has witnessed enhancements, such as LLM-based data augmentation. Nonetheless, prior research harbors two primary concerns: firstly, a lack of contemplation regarding whether the natural language generated by LLM (LLMNL) truly aligns with human natural language (HNL), a critical foundational question; secondly, an oversight that augmented data is randomly generated by LLM, implying that not all data may possess equal training value, that could impede the performance of classifiers. To address these challenges, we introduce the scaling laws to intrinsically calculate LLMNL and HNL. Through extensive experiments, we reveal slight deviations (approximately 0.2 Mandelbrot exponent) from Mandelbrot's law in LLMNL, underscore a complexity advantage in HNL, and supplement an interpretive discussion on language style. This establishes a solid foundation for LLM's expansion. Further, we introduce a novel data augmentation method for few-shot text classification, termed ZGPTDA, which leverages fuzzy computing mechanisms driven by the conformity to scaling laws to make decisions about GPT-4 augmented data. Extensive experiments, conducted in real-world scenarios, confirms the effectiveness (improving F1 of Bert and RoBerta by 7-10%) and competitiveness (surpassing recent AugGPT and GENCO methods by about 2% accuracy on DeBerta) of ZGPTDA. In addition, we reveal some interesting insights, e.g., Hilberg's law and Taylor's law can impart more benefits to text classification, etc.

## 1 INTRODUCTION

Recently, the emergence of large language models (LLMs) like GPT-4 has sparked a frenzy within the AI community, due to their revolutionary text generation capabilities. LLMs have revitalized natural language processing (NLP) by generating language/text that is extremely similar to human-generated content, such as in few-shot text classification tasks [1-10].

However, a critical issue remains: the veracity of LLM-generated natural language (LLMNL) in truly replicating human natural language (HNL) remains an enigma. Previous research has posited that the training of LLMs, designed through reinforcement learning from human feedback, naturally presupposes that the generated text aligns with HNL. However, the empirical veracity of this assumption remains largely unexplored, casting shadows of doubt over its foundational legitimacy. The first objective of this paper is to critically evaluate the rationality of this presumption, laying a theoretical groundwork for its application in related domains. Moreover, in real-world scenarios of few-shot text classification, due to the high cost of LLM itself participating in classification and its limited understanding in certain vertical fields, current research primarily relies on pre-trained language models (like the Bert family) and utilizes LLMs to generate additional training sets from original datasets to address the issue of insufficient training data [1-3, 9, 11-13]. This approach, known as data augmentation, has been proven to be an efficient and popular solution [28-32]. However, previous research has overlooked the fact that the text generated is random and may include content of insufficient value. This oversight could limit the performance of classification models.

Embarking on the quest to quantitatively dissect the LLMNL and HNL presents an intellectually daunting endeavor. Previous research has gravitated towards employing metrics such as perplexity, which is predicated on the precision of word predictions, alongside n-grams co-occurrence metrics like BLEU and ROUGE [14]. These methodologies predominantly illuminate the structural facets of language, embodying a mechanistic quantification that, while instrumental, only skims the surface of linguistic profundity. The exigency for a methodology that penetrates beyond the superficial into the quintessence of language is palpable. In this context, the scaling laws of natural language emerge as a beacon of insight. Informed by the wisdom of statistical physics [15], these laws advocate for a perception of language not as a mere aggregation of syntactic and semantic units but as a complex system. This paradigmatic shift enables an exploration of language through the prism of its microscopic constituents, such as words, to unveil the statistical regularities that govern its macroscopic phenomena—the intrinsic attributes of language [16]. A paradigmatic illustration of this wisdom is Zipf's law of word frequencies, as proven by the principle of least effort [17], which elucidates a linguistic economy, where the objective to maximize communicative efficiency while minimizing cognitive and linguistic exertion leads to a skewed distribution of word frequencies within a corpus. It is as if the lexicon's statistical distribution morphs into an effortless conduit for the transmission of complex ideas, engendering the empirically observed phenomena in Zipf's law [16].

Recent study [14] provides an initial foundation for this paper by showcasing four scaling laws: Zipf's law, Heaps' law, Ebeling's law, and Taylor's law. While this study extends into the long-range correlations through the autocorrelation function, there is a more appropriate and profound Mandelbrot-fractal, known to some extent as Mandelbrot's law [16]. This law, viewed through the prism of multifractal analysis, offers an enriched understanding [18, 20]. Also, this paper incorporates other well-known laws such as Hilberg's law, Menzerath's law, and Benford's law [22-24], broadening the scope of investigation.

In navigating the second challenge, the insights gleaned from addressing the first question catalyze a new text data augmentation methodology, termed ZGPTDA. ZGPTDA innovates upon the motivation that LLM generates text randomly, and thus not all generated texts uniformly beneficial for training purposes. Furthermore, the generated texts that are closer to human language should be deemed more appropriate because classifiers are designed to serve humans and are used in real-life situations. Therefore, ZGPTDA evaluates the suitability of these texts based on their degree of conformity to eight scaling laws, such as the goodness of fit, to identify the expected augmented instances through a decision-making process mediated by fuzzy computing mechanisms.

Formally, this paper proposes two research objectives: (1) The emergence of LLM appears to blur the language boundaries between humans and machines in NLP, yet empirical investigation into this phenomenon is largely lacking. This paper aims to introduce the scaling law to fill this gap. (2) While LLM-assisted data augmentation has facilitated progress for text classification, previous research has overlooked the fact that not all texts generated by LLM may hold the same training value. This paper seeks to propose a new data augmentation approach.

The contributions unfold as follows:

1. We introduce an innovative methodology that harnesses scaling laws to scrutinize the similarities and differences between LLM and human natural language, which provides a theoretical basis and confidence for related research.

Also, we engage with eight scaling laws as thoroughly as possible across multiple datasets, such as Mandelbrot's law. To the best of our knowledge, this represents the first such comprehensive quantitative exploration.

2. We propose a new data augmentation method for text classification, ZGPTDA, which leverages a Z-number mechanism to govern the selection of expected augmented data based on a quantified alignment with scaling laws by the corpus generated by GPT-4. ZGPTDA's robustness and competitiveness have rigorously evaluated through extensive experiments, e.g., improving F1 of Bert and RoBerta classifiers by 7-10%, and surpassing recent AugGPT and GENCO approaches by about 2% accuracy on DeBerta, etc.

3. We reveal a series of novel insights. Our findings indicate that the congruence between LLMNL and HNL is substantially established, positioning LLM-generated text as a viable supplement for NLP applications. Moreover, we critically elucidate the fractal complexity—or the relative lack thereof—within LLMNL, which is further interpreted with the aid of language style such as readability, sentiment, and semantics, offering additional understanding of LLM itself and subsequent iterations. Additionally, we demonstrate that not all LLM-generated data possess equal training value from an experimental standpoint, and scaling laws-based fuzzy operations can emerge as efficacious methods for managing such disparities. Furthermore, we ascertain that Hilberg's law and Taylor's law can impart more benefits to text classification, which is poised to inspire subsequent research within the academic community, such as prioritizing these laws in feature engineering to optimize efficiency.

## 2 RELATED WORK

### 2.1 LLM-based data augmentation

In the vanguard of contemporary AI advancements lies the deployment of LLM (e.g., GPT-4) marking a paradigm shift in NLP tasks. Moreover, the incorporation of reinforcement learning from human feedback has further refined LLM, endowing them with the capability to produce more natural and contextually appropriate responses [25-26].

Despite its monumental success, LLM is not without concerns. One critical issue is that their autoregressive nature is inherently tailored more towards generation than understanding. This distinction becomes particularly significant in specialized verticals, such as industrial safety [27], where the paucity of domain-specific training data can render LLM less effective, and there are often limitations when directly used for tasks such as hazard classification [19]. Furthermore, the substantial costs associated with the training and deployment of LLM cannot be overlooked. Given these challenges, LLM encounters difficulties when applied directly to certain NLP tasks, but rather acts as an advanced assistive tool, especially impressive for few-shot text classification [1, 3, 9, 11-13].

In scenarios where labeled data is scarce, a crucial challenge lies in training an accurate text classification model [33]. The scarcity of labeled data often stems from difficulties in acquisition due to privacy concerns or the high costs associated with manual annotation. LLM data augmentation emerges as a direct and effective strategy to mitigate this dilemma: using LLM-generated labeled texts as an augmentation of the original training dataset size [9]. This strategy addresses a critical limitation of current text data augmentation methods [33, 34]: the trade-off between fidelity and diversity. On the one hand, ensuring the correctness of the generated data's labels (fidelity) is paramount, as inaccuracies can lead to poor model performance. On the other hand, generating data that is sufficiently varied (diversity) is essential for building robust models that can generalize well to new, unseen data. Previous studies typically involve guiding LLM with prompts to rewrite each text in the original training set (*Draw*) into multiple augmented texts (*Daug*). The strategies for integrating *Daug* into classifier training are varied: (1): *Draw* and *Daug* are concatenated to form a new training dataset, which is then used to train the classifier [3]. (2): The label prompts transform the label name to natural language text, which is coordinated with the contrastive loss function to incorporate *Draw* and *Daug* into classifier training [9]. (3): The embeddings of *Daug* are averaged and merged with those of *Draw*. This combined embedding is then used for training the classifier [11]. (4): *Daug* is optimized by heuristic algorithms and then concatenated with *Draw* to complete classifier training [1]. (5): *Draw* is adjusted based on oversampling techniques before being concatenated with *Daug*. The new training set thus formed is used for classifier training [13]. Where, the choice of classifier often falls on the efficient models from the Bert family

Nevertheless, there are two primary concerns that merit careful consideration. The first concern revolves around the authenticity and consistency of *Daug* with human language. There is an uncertainty regarding whether *Daug* truly mirrors the nature of human language. Empirical experiments are essential to solidify this theoretical foundation. The second concern pertains to the potential presence of unsuitability within *Daug*. In many cases, there is a lack of strategic filtering, which means that low-quality data (that is not natural enough) might be included in the training set. This oversight can adversely affect the model's performance.

**2.2  Scaling laws of natural language**

Statistical physics reveals that the concept of scaling laws pertains to the manifestation of power-law correlations among diverse variables under the diminution of scale in complex systems. These laws epitomize an emergent phenomenon, symbolizing an incisive traversal beyond the superficial manifestations of a system and capturing its intrinsic essence. Within the natural language, the structural anatomy of text evolves as a complex system [15], an amalgamation of discrete constituents like words. The dynamism of language is derived from the intricate synthesis of a limited repertoire of elements, culminating in a boundlessly diverse spectrum of textual expressions. Although isolating insights into the quintessence of language from its multifaceted factors presents a formidable challenge, the pervasive influence of scaling laws offers a dimensional paradigm for elucidating the nature of language system [16].

The endeavor to quantitatively distill the essence of natural language through scaling laws, particularly in juxtaposing machine-generated and human language, transcends nascent idea, which has witnessed concrete success. For instance, the study employs Zipf's law and Heaps' law in a rigorous assessment of texts synthesized by LSTM [35]. Further exemplifying this trend is the study [14], which embarks on analysis of the texts generated by generative adversarial networks and recurrent neural networks, which is anchored on: Zipf's law, Heaps' law, Ebeling's law, Taylor's law, and the concept of the long-range correlation.

Nonetheless, these studies, while pioneering, are not without their limitations in both breadth and depth. For instance, the application of the autocorrelation function to delineate Long-range correlations, though insightful, presents a somewhat rudimentary and constrained perspective. A more sophisticated and encompassing alternative is the Mandelbrot-fractal, also known to some extent as Mandelbrot's law [16, 18], predicated on multifractal analysis [19-20]. This law unveils the fractal-like self-similarity of language, suggesting that microcosmic segments of a text can, in certain respects, mirror and encapsulate the overarching narrative or meaning, thereby illuminating the pervasive interconnections within linguistic structures.

Moreover, this paper inquiry extends its purview to other eminent linguistic scaling laws, namely Hilberg's law, Menzerath's law, and Benford's law. Each of these laws offers distinct yet complementary insights: Hilberg's law probes into the scaling of information density vis-à-vis text length, serving as an extension or reinterpretation of Claude Shannon's foundational work in information theory [24]. Menzerath's law elucidates the organizational structure of language units and proposes a general trend where the length of a linguistic structure is inversely related to the size of its components [23]. Benford's law, focuses on the distribution of numbers within textual data. It predicts first digit is 1 about 30% of the time, and the likelihood of each subsequent number as the first digit decreases logarithmically [22].

## 3 EVALUATION FOR NATURAL LANGUAGE

### 3.1 Scaling laws

We explore Zipf's law, Heaps' law, Taylor's law, Hilberg's law, Ebeling's law, Menzerath's law, Benford's law, and Mandelbrot's law, with details as follows.

*3.1.1 Zipf's law*

Zipf's law, delineates that word frequency *f(r)*, is a function of its rank, *r*:

$$f(r) \propto r^{\alpha} \quad (1)$$

This law reflects a deeper underlying principle of economy in human language – a preference for minimizing effort, which manifests in a small set of words being used disproportionately often. This principle of least effort applies not only in linguistic communication but also in a wide range of human behaviors [17]. From a systemic perspective, Zipf's law is connected to the tenets of self-organization, where elementary and repetitive interactions at the micro level — exemplified by the selection of words— culminate in the emergence of complex and structured patterns at the macro level, as evidenced by the overarching distribution of word frequencies.

*3.1.2 Heaps' law*

Heaps' law articulates a correlation between the magnitude of a textual corpus and its vocabulary diversity. Define *n* as the extent (vocabulary count) of the text, and *v(n)* as the corresponding magnitude of its vocabulary, the law is thereby encapsulated in Eq.2.

$$v(n) \propto n^{\beta} \quad (2)$$

Where, *β*, a parameter greater than zero, signifies that with the expansion of the corpus, the proliferation of unique vocabularies escalates, yet this incrementation manifests at a progressively attenuated velocity. This principle resonates with an understanding: as one delves deeper into literature pertaining to a specific subject, the incidence of encountering new information diminishes.

*3.1.3 Taylor's law*

Taylor's law posits that the variance in the frequency of words adheres to a consistent power-law pattern, regardless of the average density. For a given fragment size in a text, the occurrence t of a unique word w is noted, and its average, $\rho_w$, along with its standard deviation, $\sigma_w$, are computed. The Taylor exponent $\zeta$ elucidates the following scaling relationship.

$$\sigma_w(t) \propto \rho_w^{\zeta} \quad (3)$$

*3.1.4 Hilberg's law*

Hilberg's law addresses the dynamics of entropy of natural language, as it scales with the expanding breadth of a textual corpus. Specifically, the entropy $\gamma(\mu)$ of word blocks μ, aside from a contentious constant and linear terms, exhibits a power-law relationship with the *μ*, where the exponent is less than one.

$$\gamma(\mu) \propto \mu^{\phi} \quad (4)$$

This relationship suggests that with the elongation of textual sequences, there is an escalation in information content, yet this augmentation occurs at a progressively decelerating rate. In other words, larger volume of text does not necessarily become proportionally more complex or information-rich. This observation can be interpreted as an embodiment of natural language's intrinsic redundancy; a significant portion of the material conveyed in more extended texts tends to be either repetitive or predictable.

*3.1.5 Ebeling's Law*

Ebeling's law reflects the interplay between stochasticity and variability within linguistic constructs. It articulates a power-law relationship, correlating the lengths of textual subsequences with the variance in character occurrences within these subsequences. From a defined ensemble of elements, W, comprising individual characters c, the function m(c, u) quantifies the variance (Var) of frequency of c within W of a designated length u. Then, there is:

$$m(u) = \sum_W Var[m(c,u)] \propto u^\eta \quad (5)$$

*3.1.6 Menzerath's law*

Menzerath's law pertains to the linguistic unit organization, encapsulating a counterintuitive principle: "The greater the whole, the smaller its parts." Where an increase in sentence length $\vartheta(l)$ typically coincides with a reduction in the average length (letter count) $l$ of its constituent words.

$$\vartheta(l) \propto l^\varphi \quad (6)$$

It reflects a pivotal insight of natural language, highlighting the influence of processing efficiency and usability in the evolution and formation of linguistic structures, where the larger linguistic constructs are counterbalanced by a tendency towards conciseness and simplicity in individual components, embodying a deep-rooted optimization in human communicative practices.

*3.1.7 Benford's law*

Benford's law delineates a log-linear distribution in the prevalence *f(d)* of digits (*d*) 1 through 9 within numbers (e.g., sentence length variations) in natural language:

$$f(d) = \log_{10}(1 + \frac{1}{d})$$

A recent study has expanded the scope of this statistical principle [36], aligning more closely with a Gamma distribution characterized by the parameter κ approaching zero, thus manifesting a quasi-scaling law:

$$f(d) \propto e^{-\kappa d} d^{\omega - 1} \quad (7)$$

It acts as a diagnostic instruction for assessing the naturalness of linguistic phenomena.

*3.1.8 Mandelbrot's law*

Mandelbrot's law sketches the multifractal in natural language. Given a text, Bert vectorizes and averages it to $V = \{v_1, v_2, …, v_n\}$, then, its profile sequence *L(j)* is formed by:

$$L(j) \equiv \sum_{k=1}^{j}[V(k) - avg(V)], \quad j = 1, 2, ..., n$$

*L(j)* is divided into $2M_s$ non-overlapping windows *v* of size *s* through a bidirectional sliding process, thereby obtaining the detrended variance.

$$F^2(v, s) = \frac{1}{s}\{L((v-1)s + k) - P_v^{(m)}(k)\}$$

Where the m-order polynomial $P_v^{(m)}$ achieves detrending. And, the fluctuation function is calculated:

$$F_q(s) = \{\frac{1}{2M_s}\sum_{v=1}^{2M_s}[F^2(v,s)^{q/2}]\}^{1/q}$$

As the order *q* changes, the Long-range correlation is emerged by:

$$F_q(s) \propto s^{h(q)} \quad (8)$$

Where, *h(q)* is conceptualized as a function of *q*, which unfolds self-similarity at divergent scales of text. This suggests that there are the ubiquitous interconnections within language, where smaller units of language, at times, do not merely contribute to the aggregate meaning but often mirror and distill the overarching narrative or thematic essence of the entire corpus in a more concentrated and revealing manner.

### 3.2 Experiment setting

Three datasets are anchored, each containing texts produced by both GPT-3.5 and human on the same prompt. All LLMNL and HNL in each dataset are integrated separately to better conduct experiments. Table 1 shows some statistical information.

1. Cgtd [37]: It is based on TOEFL essays.
2. Cheat [37]: It is compiled from a variety of sources: Quora interpretation questions, SQUAD 2.0 dataset, and topics covered in CNN news articles.
3. HC3 [38]: It encompasses five distinct topics within the realms of medicine and finance.

Table 1: Data information.

| Dataset | Char count | | Word count | | Sentence count | |
|---|---|---|---|---|---|---|
| | Human | LLM | Human | LLM | Human | LLM |
| Cgtd | 272073 | 252990 | 53063 | 45750 | 2465 | 1717 |
| Cheat | 16690110 | 35909844 | 2443101 | 5114311 | 117989 | 246762 |
| HC3 | 39851408 | 27185722 | 7778270 | 4629238 | 409032 | 219368 |

The parameters of scaling laws are optimally determined through regression analysis [14]. The goodness of fit is measured by $R^2$, see Eq.9, where $p$ and $q$ are observed and fitted values, and $N$ is the size of the dataset. $R^2 \in [0, 1]$, and $R^2 > 0.9$ signifies a strong adherence.

$$R^2 = 1 - \frac{\sum_{i=1}^{N}(p_i - q_i)^2}{\sum_{i=1}^{N}(p_i - avg(p_i))^2} \quad (9)$$

Although $R^2$ is highly informative, considering real-world data rarely strictly follows laws, we have included additional metrics. The Kullback-Leibler divergence (KL) test is a widely utilized measure to gauge the similarity between two distributions by quantifying the difference between their predicted and actual distributions, see Eq.10. $KL \in [0, \infty)$, a smaller KL signifies a higher degree of consistency, and typically, $KL < 0.5$ is considered acceptable. Similarly, the Jensen-Shannon divergence (JS) test, another extensively utilized measure, see Eq.11. $JS \in [0, 1]$, a lower value indicates a higher degree of concordance, and $JS < 0.2$ is often deemed appropriate. The final metric, the mean absolute percentage error (MAPE), quantifies the discrepancy between fitted and actual values, See Eq.12. $MAPE \in [0, +\infty)$, and $MAPE < 0.5$ is generally considered reasonable.

$$KL = \sum_{i=1}^{N} p_i \log \frac{p_i}{q_i} \quad (10)$$

$$JS = \frac{1}{2}\sum_{i=1}^{N} p_i \log \frac{p_i}{p_i + q_i} + \frac{1}{2}\sum_{i=1}^{N} q_i \log \frac{q_i}{p_i + q_i} + \log 2 \quad (11)$$

$$MAPE = \frac{1}{N}\sum_{i=1}^{N} \left| \frac{q_i - p_i}{p_i} \right| \times (100\%) \quad (12)$$

### 3.3 Results

Table 2 substantiates that LLM, parallels to human, also successfully exhibits scaling laws, as evidenced through four key indicators, displaying reasonableness and appropriateness (with the exception of minor imperfections in MAPE for certain laws). Remarkably, all $R^2$ are consistently above 0.9, and even surpass 0.99 in Heaps' law and Mandelbrot's law across three datasets. Also, the minimal values of KL and JS divergences (e.g., as low as 0.001 in Mandelbrot's law) robustly corroborate the alignment between the scaling laws and the true distribution manifested by LLM. This comprehensive evidence eloquently attests to LLM's authentic linguistic behavior, mirroring the patterns observed in human.

Fig.1 elucidates an extraordinary congruence between the emergent scaling laws in the language outputs of LLM and human across various datasets, manifesting uniformity in trends. Specifically, in each {LLM, human} pair, (1): Discrepancies in Zipf exponent α are constrained below 0.03 (illustrated by -0.7936 vs. -0.7762 on HC3), indicative of a nearly equivalent commitment to the principle of least effort. (2): Heaps exponents β exhibit close proximity (as seen with 0.8542 vs. 0.8094 on Cheat), reflecting a parallel escalation in the difficulty of extracting new information (represented as unique words) as the corpus size expands. (3): Ebeling exponents η are strikingly similar (e.g., 1.3150 vs. 1.3022 on Cgtd), elucidating a comparable equilibrium attained between randomness and order within both modalities. (4): Hilberg exponents φ maintain a consistent profile, marginally exceeding 0.1 (e.g., 0.1281 vs. 0.1244 on Cgtd), signifying that their information entropy conveyed follows a sub-linear trajectory, reflective of substantial linguistic redundancy. (5): Menzerath exponents φ hover around -1, alluding to their potentially uniform structural organization, where the relationship between the magnitude of the whole and the diminution of its parts follows a log-linear pattern. (6): Taylor exponents ζ are closely aligned (e.g., 0.6821 vs. 0.6338 on Cgtd), indicating a parallel increase in variability with their heightened average density or richness. (7): Benford exponents κ and ω exhibit certain differences, primarily due to the predominance of the digits 1 or 2. This phenomenon may be ascribed to the constrained span or magnitude of sentence lengths, thereby limiting numerical representation within the datasets.

Table 2: Results of natural language evaluation.

| Scaling law | Dataset | $R^2$ | | KL | | JS | | MAPE | |
|---|---|---|---|---|---|---|---|---|---|
| | | LLM | Human | LLM | Human | LLM | Human | LLM | Human |
| Zipf's law | Cgtd | 0.9391 | 0.9548 | 0.0491 | 0.0521 | 0.0175 | 0.0190 | 0.6654 | 0.7468 |
| | Cheat | 0.9799 | 0.9882 | 0.0136 | 0.0088 | 0.0048 | 0.0031 | 0.3023 | 0.1678 |
| | HC3 | 0.9447 | 0.9584 | 0.0757 | 0.0719 | 0.0271 | 0.0262 | 0.9994 | 1.0031 |
| Heaps' law | Cgtd | 0.9980 | 0.9978 | 0.0006 | 0.0004 | 0.0002 | 0.0002 | 0.0373 | 0.0398 |
| | Cheat | 0.9997 | 0.9978 | 0.0001 | 0.0004 | 0.0001 | 0.0001 | 0.0132 | 0.0366 |
| | HC3 | 0.9983 | 0.9994 | 0.0005 | 0.0001 | 0.0002 | 0.0001 | 0.0365 | 0.0173 |
| Mandelbrot's law | Cgtd | 0.9923 | 0.9429 | 0.0001 | 0.0002 | 0.0001 | 0.0001 | 0.0055 | 0.0158 |
| | Cheat | 0.9986 | 0.9880 | 0.0001 | 0.0001 | 0.0001 | 0.0001 | 0.0017 | 0.0064 |
| | HC3 | 0.9910 | 0.9785 | 0.0001 | 0.0001 | 0.0001 | 0.0001 | 0.0046 | 0.0099 |
| Taylor's law | Cgtd | 0.9304 | 0.9265 | 0.0481 | 0.0393 | 0.0169 | 0.0138 | 0.0226 | 0.1912 |
| | Cheat | 0.9391 | 0.9429 | 0.0290 | 0.0284 | 0.0100 | 0.0097 | 0.0179 | 0.0164 |
| | HC3 | 0.9134 | 0.9305 | 0.0380 | 0.0318 | 0.0130 | 0.0108 | 0.1045 | 0.0430 |
| Hilberg's law | Cgtd | 0.9852 | 0.9724 | 0.0002 | 0.0003 | 0.0001 | 0.0001 | 0.0109 | 0.0144 |
| | Cheat | 0.9708 | 0.9786 | 0.0003 | 0.0003 | 0.0001 | 0.0001 | 0.0178 | 0.0148 |
| | HC3 | 0.9608 | 0.9663 | 0.0004 | 0.0004 | 0.0002 | 0.0002 | 0.0213 | 0.0206 |
| Ebeling's law | Cgtd | 0.9994 | 0.9997 | 0.0021 | 0.0010 | 0.0007 | 0.0004 | 0.1204 | 0.0920 |
| | Cheat | 0.9897 | 0.9901 | 0.0062 | 0.0030 | 0.0023 | 0.0011 | 0.1730 | 0.0910 |
| | HC3 | 0.9916 | 0.9941 | 0.0194 | 0.0059 | 0.0065 | 0.0021 | 0.3034 | 0.1746 |
| Menzerath's law | Cgtd | 0.9950 | 0.9983 | 0.0017 | 0.0031 | 0.0006 | 0.0011 | 0.0703 | 0.1302 |
| | Cheat | 0.9678 | 0.9673 | 0.0105 | 0.0107 | 0.0038 | 0.0038 | 0.2039 | 0.2066 |
| | HC3 | 0.9994 | 0.9980 | 0.0012 | 0.0076 | 0.0004 | 0.0027 | 0.0743 | 0.1949 |
| Benford's law | Cgtd | 0.9992 | 0.9880 | 0.0366 | 0.3591 | 0.0054 | 0.0412 | 0.4252 | 0.5211 |
| | Cheat | 0.9795 | 0.9953 | 0.0678 | 0.0307 | 0.0172 | 0.0075 | 0.5570 | 0.3351 |

| | HC3 | 0.9735 | 0.9815 | 0.0609 | 0.0553 | 0.0159 | 0.0151 | 0.4188 | 0.2991 |

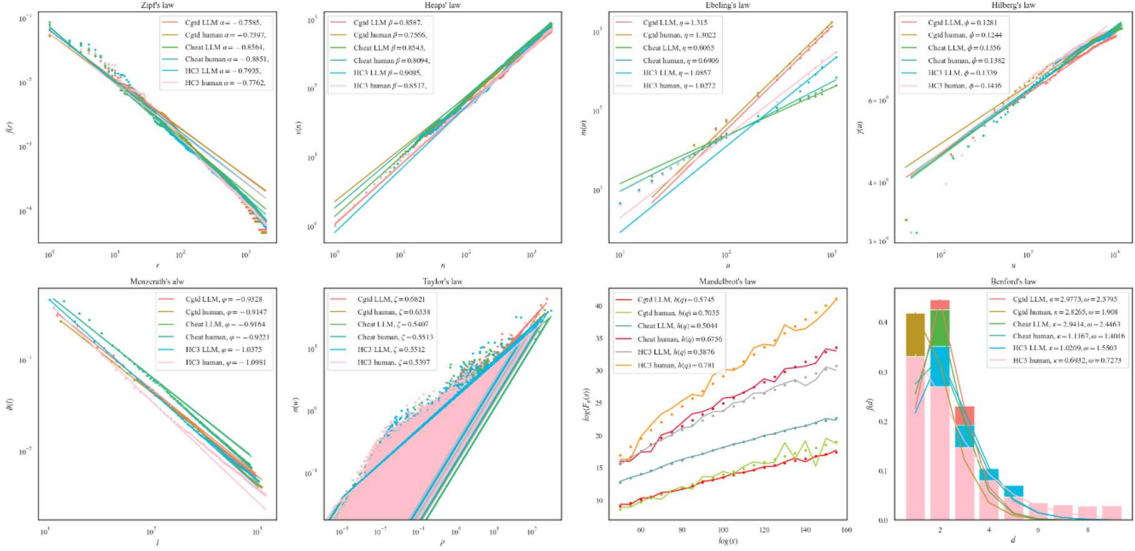

Fig.1: Emergence of scaling laws between LLM and human natural language.

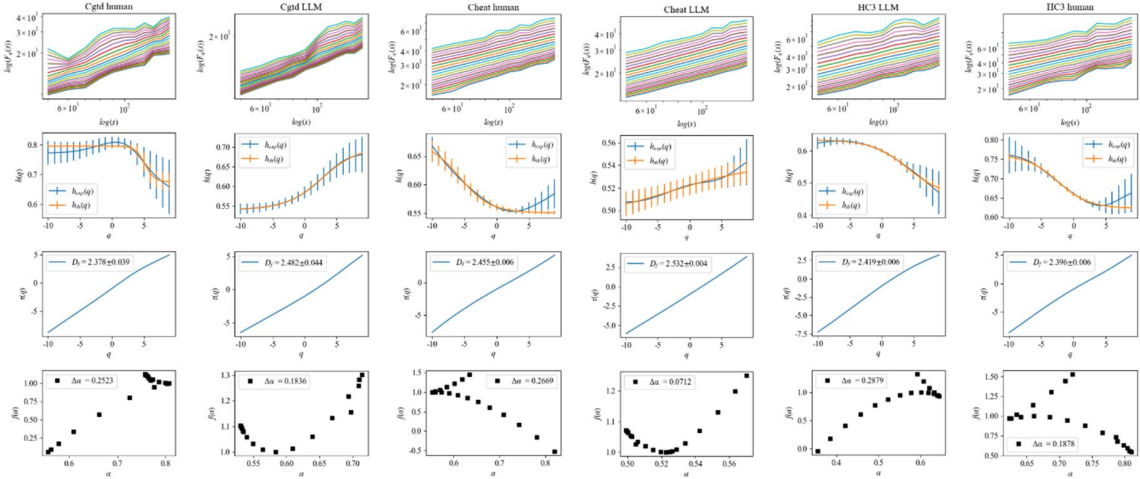

Fig.2: Multifractal analysis between LLMNL and HNL.

Intriguingly, despite Mandelbrot exponents *h(q)*-based scales universally exhibiting a trend of gradual ascension with increasing window size *s*, a distinct pattern emerges: the *h(q)* values for NHL across all datasets invariably surpass those of LLM. To locate this divergence, a more granular exploration of the multifractal behaviors is driven by varying the order *q*. The Boltzmann machine serves Eq.8 to ascertain *h(q)* from the fluctuation function $F_q(s)$, with *q* = [-10, 10]. Fig.2 row 1 reveals a parallelism in scale-based fluctuations between LLM and human languages, a convergence further echoed in their *h(q)* values predominantly exceeding 0.5 (see Fig.2, row 2). This pattern indicates a pronounced presence of long-range correlations within both language frameworks. Further delineation of the scaling of moments of the distribution is achieved by examining $\tau(q)$, calculated from $q \times h(q) - D_f$, where $D_f$ denotes the fractal dimension. This reaffirms the alignment in fractal attributes between LLMNL and HNL, see Fig.2, row 3. The investigation extends to characterizing the Hölder exponent $\Delta\alpha = \alpha_{max} - \alpha_{min}$, where $\alpha$ is defined as $h(q) + qh'(q)$ and $f(\alpha) = q \times [\alpha - h(q)] + 1$ (see Fig.2, row 4), which, ultimately, from the degree of complexity, unveils a nuanced disparity, especially highlighted in Cheat. Here, the complexity of LLMNL, with $\Delta\alpha = 0.0712$, does not quite attain that in HNL, which stands at $\Delta\alpha = 0.2669$.

Conclusively, the rigorous examination of eight distinct scaling laws has demonstrated that, despite some minor differences in the degree of fractal-complexity, LLMNL exhibits a remarkable congruence and parity with HNL across multiple dimensions. This discovery underpins a substantial foundation for subsequent NLP research endeavors, offering a reinforced confidence in employing LLMNL, especially for data augmentation of text classification. The ensuing section presents a case study to elucidate an innovative data augmentation technique, rooted in the synergy of LLMs and scaling laws, which promises to significantly bolster the efficacy and scope of NLP applications.

## 4 EVALUATION FOR NATURAL LANGUAGE

This paper delineates an innovative paradigm for data augmentation, designated as ZGPTDA. The underlying premise of ZGPTDA is an acknowledgment of the inherent randomness in the texts generated by LLM, implying that the texts exhibit varying degrees of suitability. In other words, not all instances hold equal value for training purposes. The crux of ZGPTDA lies in evaluating these instances based on eight distinct scaling laws, wherein the degree of emergence with these scaling laws quantitatively defines each instance's "suitability". Instances exhibiting higher degrees of suitability are deemed more representative of human language and are thus considered more authentic. Consequently, it is advantageous to selectively incorporate these more suitable instances into the training process, thereby enhancing the overall performance of the final model. The measurement of suitability is facilitated by metrics $R^2$, JS, KL, and MAPE. To further comprehend this framework, one can draw parallels from real-world applications, e.g., "scaling laws help determine the healthiness of food," which analyzes the degree of healthiness of food based on the extent to which its nutritional components deviate from scaling laws [39].

In light of the metrics presenting a state of ambiguity regarding suitability (for instance, roughly suitable or highly suitable), and considering the differences among these metrics, the selection of suitable instances poses substantial challenges. To address this, ZGPTDA incorporates the fuzzy logic, utilizing the widely acclaimed Z-number theory [40] that has seen substantial success in numerous decision-making processes. The Z-number consists of a pair of fuzzy components, in this paper, the first component represents a fuzzy constraint, and the second is a measure of reliability. This dual-component structure facilitates an aggregation framework, enabling the synthesis of these metrics, concurrently considering both their magnitudes and associated reliability, to ascertain the overall suitability of instances.

### 4.1 ZGPTDA

Initially, for each text in the raw dataset $D_{raw}$, ZGPTDA employs a straightforward prompt to guide GPT-4 in generating some augmented instances, such as "You are a famous expert in the field of engineering. Based on your understanding, restate the text in $n$ sentences". Here, $n = 10$ in this paper. Hence, an instance set $T = \{t_1, t_2, …, t_n\}$ is obtained. Each instance $t_k$ is then subjected to fitting against eight scaling laws $L_i$, $i = 1, …, 8$. This process forms value sets of their metrics: $PL_i = \{PL_1, PL_2, …, PL_8\}$, where $PL_i = \{R_i^2, KL_i, JS_i, MAPE_i\}$.

Subsequently, a Z-number comprising components $A$ and $B$, is constructed. Here, the component $A$ signifies the extent to which each $t_k$ adheres to the metrics based on each law, denoted as $A = (A_{R^2,i}, A_{KL,i}, A_{JS,i}, A_{MAPE,i})$, where each element is conceptualized as a fuzzy set. These sets are defined by a membership function $\mu$, adeptly mapping the spectrum of measurement values onto a continuum of membership values ranging from 0 to 1. These values, in accordance with their assigned range, quantitatively reflect the extent to which $t_k$ conforms to the conditions delineated by each fuzzy set, exemplified by varying degrees such as "High", "Medium", and "Low". Central to this manipulation is the implementation of a triangular membership function $trimf(x)$, delineated by a triad of points: $(a, b, c)$, with each point's value determined through the experience of three experts. Specifically, in each law:

For $R^2$:

$$\mu_{1-R^2, Low}(x) = trimf(x; [0, 0.05, 0.1])$$
$$\mu_{1-R^2, Medium}(x) = trimf(x; [0.1, 0.15, 0.2])$$
$$\mu_{1-R^2, High}(x) = trimf(x; [0.2, 0.6, 1])$$

For KL:

$$\mu_{KL, Low}(x) = trimf(x; [0, 0.1, 0.2])$$
$$\mu_{KL, Medium}(x) = trimf(x; [0.2, 0.35, 0.5])$$
$$\mu_{KL, High}(x) = trimf(x; [0.5, 0.75, 1])$$

For JS:

$$\mu_{JS, Low}(x) = trimf(x; [0, 0.05, 0.1])$$
$$\mu_{JS, Medium}(x) = trimf(x; [0.1, 0.15, 0.2])$$
$$\mu_{JS, High}(x) = trimf(x; [0.2, 0.6, 1])$$

For MAPE:

$$\mu_{KL, Low}(x) = trimf(x; [0, 0.1, 0.2])$$
$$\mu_{KL, Medium}(x) = trimf(x; [0.2, 0.35, 0.5])$$
$$\mu_{KL, High}(x) = trimf(x; [0.5, 0.75, 1])$$

Where,

$$trimf(x) = \begin{cases} (x-a)/(b-a), & x \in [a,b] \\ (c-x)/(c-b), & x \in [b,c] \\ 0, & x \in others \end{cases}$$

On the other hand, the $B$ Component embodies the reliability of the statement made by $A$. Considering that $A$ comes from a variety of sources, we construct $B_i = \sigma(A_i)$, where $\sigma(x)$ is the standard deviation operation.

The aggregated metric assessment $A_t$ and total reliability measure $B_t$ for each text are derived through a weighted amalgamation of the individual metrics' fuzzy sets:

$$A_t = \frac{1}{8} \sum_{i=1}^{8} \| WA_i^T \|; \quad B_t = \frac{1}{8} \sum_{i=1}^{8} \| WB_i^T \|$$

Where, $W$ is weight matrix $[W_{R^2}, W_{KL}, W_{JS}, W_{MAPE}]$ with values of 0.1, 0.2, 0.2, and 0.5, empirically.

Then, ZGPTDA leverages fuzzy inference rules to map the $A_t$ and $B_t$ to a suitability decision. This process can be represented as a mapping based the Mamdani method [41], e.g.:

$$\min(\mu_H(A_t), \max(\mu_L(B_t), \mu_M(B_t), \mu_H(B_t))) \to \mu_H(S')$$
$$\min(\mu_M(A_t), \max(\mu_M(B_t), \mu_H(B_t))) \to \mu_M(S')$$
$$\min(\mu_L(A_t), \mu_H(B_t)) \to \max(\mu_L(S), \mu_M(S'))$$
$$\min(\mu_L(A_t), \max(\mu_L(B_t), \mu_M(B_t))) \to \mu_L(S')$$

Where the membership functions for "High", "Medium", and "Low" are denoted as $H$, $M$ and $L$, $S'$ is the non-suitability score. Thus, the Suitability score $S$ can be harvested by the Centroid defuzzification method, $S = 1 - \Phi(S')$:

$$\Phi(x) = \frac{\int_x x \cdot f(x) dx}{\int_x f(x) dx}$$

Ultimately, ZGPTDA processes the instance set $T$, thereby quantifying $S$ for each instance. These instances are ranked in descending order based on their $S$ values, facilitating the strategic selection of the suitable augmented ones. For the sake of simplicity, the top 50% of $T$, representing a quintuple augmentation, are extracted to form the augmented instance set, $D_{aug}$. $D_{aug}$ is concatenated with $D_{raw}$ to form a new training-set, for training classification models.

**4.2 Experiment setting**

The application case chosen for this study is the classification of hazard events in the context of Hazard and Operability paradigm, a decision motivated by its practical significance and industry value [27, 42]. This framework is recognized as a paragon in the safety of industry, dedicated to identifying and assessing hazard events in various industrial and societal systems, with these events being documented in text form. The classification of such hazard events is not merely an academic exercise but a pivotal element in guiding comprehensive strategic planning and informed decision-making within the related sectors. This extends to facilitating rigorous safety assessments and hazard remediation by professionals, including experts and engineers [21, 44]. There are three distinct hazard event datasets, themed around severity (encompassing five levels of hazard events, stratified from level#1 to level#5), possibility (consisting 5 levels from level#1 to level#5), and risk (spanning 4 levels from level#1 to level#4) [19-20]. A random selection of 50 samples from each training set forms the basis of our experimental inquiry. This aims to rigorously evaluate the efficacy of ZGPTDA in scenarios with limited data availability, underscoring its potential in addressing the challenges of few samples' classification.

Various prevalent and effective methods are employed for competitive experiments: (1) ENDA [43], which involves adding minor noises into original data to create new samples. (2) EDA [34], which applies random word replacement, insertion, swapping, and deletion. (3) SCDA [33], which simulates typographical errors by substituting randomly chosen characters with adjacent characters. (4) Ziteg [21], which utilizes a custom GPT2 fine-tuned on the hazard event corpus to generate data. (5) AugGPT [3], which rephrases each sentence in training set as multiple samples that are conceptually similar but semantically different. GENCO [9], which enhances the semantic embedding of instances and improves the mapping to relevant labels.

In addition, the following experiments are also conducted: (1) ZGPTDA(random): randomly selecting $D_{aug}$ from ZGPTDA's $T$. (2) ChatEmb, incorporating ZGPTDA's $D_{raw}$ into $D_{aug}$ as average embeddings [11]. (3) ChatEntro, replacing Z-number with entropy gain method for competing on $D_{aug}$ from ZGPTDA's $T$ [1]. (4): The effectiveness of individual scaling laws is assessed by monitoring the alterations in ZGPTDA's performance subsequent to the sequential removal of each law. (5): The impact of each metric is evaluated. Note that, during each experiment, the only aspect that has changed is the variable being investigated, while all other components remain unchanged.

Regarding the trial models, previous studies have widely embraced and favored Bert families, reflecting a balance between cost and efficiency. Hence, Bert, RoBerta, AlBert, and DeBerta have adopted.

Additionally, all methods involve the generation of fivefold augmented instances. To ensure robustness and reliability of results, each experiment is repeated five times to obtain an average value as the result, with accuracy (Acc%) and F1-score (F1%) serving as the performance metrics.

## 4.3 Results

Table 3 presents examples generated by ZGPTDA, based on a hazard event associated with heat exchangers. This event is characterized by severity, possibility, and risk levels of {3 / 5 / 3}, respectively, and is detailed in the accompanying text conten. It is evident that the samples generated by ZGPTDA closely mirror the form of the original data, without introducing errors in writing or resorting to simple synonym substitutions, etc.

> *"The heat exchangers E0102A/B are experiencing low liquid levels. The replenishment pumps P0101A/B are malfunctioning or experiencing fluctuations in power supply. The heat exchangers E0102A/B have insufficient cooling in the tube side for hydrogen, leading to overheating in downstream processes and the PSA section. This overheating can cause hydrogen to leak from flange seals, resulting in flash fires with significant impact on personnel safety, property, and corporate reputation, and minor environmental effects. It is recommended to have a backup pump or dual power supply circuits for the plant. For E0102A/B shell side, E0103, E0104 tube side, and V0104, a staged cooling design is advised to prevent equipment overheating and flange seal leaks, which could lead to hydrogen explosions, during operational fluctuations. A SIL rank analysis is also recommended."*

Table 3: ZGPTDA-based instance cases.

|  | Content | Suitability |
|---|---|---|
| case#1 | Heat exchangers E0102A/B are facing issues with low levels of liquid. The water refill pumps, P0101A/B, are either not working properly or have unstable electricity supply. Due to inadequate hydrogen cooling in the tubes of E0102A/B, temperatures are rising in the processes that follow and in the PSA section, which may cause hydrogen to escape through the flange seals, potentially causing explosive incidents. This poses serious risks to the safety of personnel, damages property, affects the company's reputation, and has slight environmental consequences. It's suggested to use an alternate pump or set up a dual electrical supply for the facility. A gradual cooling system is recommended for the shell side of E0102A/B, as well as for the tube sides of E0103 and E0104 and V0104, to avoid overheating of equipment and seal leaks under operational changes, which could trigger hydrogen explosions. Conducting a SIL analysis is also advised. | 0.6926 |
| case#2 | The liquid levels in the heat exchangers E0102A/B are critically low. The water booster pumps P0101A/B are either breaking down or facing erratic power supply issues. There's a problem with the cooling system on the hydrogen tube side of the heat exchangers E0102A/B, causing excessive heat in the following stages and the PSA unit. Such high temperatures can lead to hydrogen escaping through the sealings of the flanges, sparking sudden fires that seriously endanger workers, damage assets, tarnish the company's image, and have slight impacts on the environment. It's suggested to install a reserve pump or arrange for a redundant electrical supply system for the facility. To mitigate the risk of equipment overheating and seal leakage, which might set off hydrogen blasts during shifts in operation, a sequential cooling setup is recommended for the shell side of E0102A/B, as well as for the tube sides of E0103, E0104, and V0104. Undertaking an analysis for Safety Integrity Level (SIL) is also proposed. | 0.6745 |
| case#3 | Low fluid levels are plaguing the E0102A/B heat exchangers, while the auxiliary pumps P0101A/B are either faulty or subjected to unstable electricity. The cooling capacity for hydrogen within the tubes of these exchangers is inadequate, causing temperature spikes in the subsequent processes and the PSA area. These spikes might lead to hydrogen leakage from the flange seals, igniting immediate blazes that pose a substantial risk to human safety, inflict property damage, damage the company's reputation, and slightly harm the environment. The recommendation is to implement a standby pump or to ensure a dual electricity supply for the installation. A tiered cooling approach is advocated for the shell side of E0102A/B, and the tube sides of E0103, E0104, and V0104, to avert the potential of equipment overheating and flange seal breaches, which may prompt hydrogen detonations when operations are unstable. A Safety Integrity Level (SIL) evaluation is advised. | 0.6628 |

Table 4: Experiment results across test set (test) and validation set (val).

| model | Severity | | | | Risk | | | | Possibility | | | |
|---|---|---|---|---|---|---|---|---|---|---|---|---|
| | test | | val | | test | | val | | test | | val | |
| | Acc | F1 | Acc | F1 | Acc | F1 | Acc | F1 | Acc | F1 | Acc | F1 |
| Bert | 58.91 | 55.14 | 59.44 | 55.13 | 51.72 | 48.92 | 53.67 | 50.34 | 57.72 | 56.97 | 59.31 | 59.02 |
| + ENDA | 62.51 | 59.67 | 63.63 | 58.66 | 55.37 | 51.35 | 57.56 | 52.87 | 60.57 | 62.35 | 62.24 | 62.54 |
| + SCDA | 64.26 | 60.94 | 65.45 | 60.55 | 56.69 | 53.31 | 58.10 | 54.63 | 63.26 | 64.88 | 63.87 | 64.91 |
| + EDA | 64.94 | 59.92 | 64.46 | 60.03 | 56.15 | 53.44 | 58.85 | 54.19 | 63.24 | 63.18 | 63.88 | 62.98 |
| + Ziteg | 65.14 | 61.08 | 65.59 | 60.78 | 56.29 | 54.42 | 59.79 | 54.61 | 64.63 | 63.27 | 64.48 | 63.17 |
| + AugGPT | 66.72 | 62.27 | 67.35 | 62.61 | 57.82 | 56.08 | 61.27 | 57.02 | 65.77 | 65.14 | 65.60 | 65.00 |
| + GENCO | 66.59 | 62.14 | 67.34 | 62.47 | 57.74 | 56.00 | 61.19 | 57.02 | 65.68 | 65.13 | 65.55 | 64.90 |
| + ZGPTDA(random) | 66.91 | 62.60 | 66.61 | 61.80 | 57.38 | 55.74 | 61.36 | 57.52 | 66.49 | 65.45 | 65.71 | 65.01 |
| + ChatEmb | 66.59 | 62.28 | 67.57 | 63.77 | 57.70 | 55.83 | 60.73 | 57.74 | 66.38 | 64.95 | 64.98 | 66.56 |
| + ChatEntro | 67.22 | 63.50 | 68.12 | 64.18 | 58.04 | 57.17 | 62.06 | 58.36 | 67.13 | 66.04 | 65.98 | 66.96 |
| + ZGPTDA | 67.91 | 64.37 | 68.85 | 64.99 | 59.03 | 57.63 | 62.83 | 59.12 | 67.61 | 66.92 | 66.83 | 67.76 |
| RoBerta | 58.75 | 55.16 | 60.97 | 56.86 | 52.14 | 49.03 | 54.85 | 50.10 | 59.28 | 56.18 | 59.37 | 58.88 |
| + ENDA | 61.10 | 57.39 | 62.73 | 59.54 | 54.57 | 51.67 | 57.80 | 51.19 | 63.26 | 59.17 | 63.59 | 62.91 |
| + SCDA | 62.01 | 59.82 | 64.50 | 63.12 | 56.01 | 55.15 | 60.33 | 55.12 | 65.26 | 61.17 | 65.42 | 64.38 |
| + EDA | 62.67 | 57.86 | 64.53 | 60.86 | 55.80 | 52.84 | 58.85 | 53.62 | 64.52 | 60.47 | 64.66 | 63.97 |
| + Ziteg | 63.82 | 58.02 | 64.60 | 60.94 | 56.40 | 52.99 | 58.85 | 53.89 | 64.63 | 61.28 | 64.67 | 64.78 |
| + AugGPT | 64.90 | 59.23 | 65.87 | 61.96 | 57.79 | 54.64 | 60.54 | 55.47 | 65.83 | 62.71 | 66.86 | 66.11 |
| + GENCO | 64.92 | 59.32 | 65.90 | 61.98 | 57.85 | 54.69 | 60.60 | 55.58 | 65.86 | 62.74 | 66.92 | 66.16 |
| + ZGPTDA(random) | 65.25 | 60.47 | 67.40 | 62.62 | 58.03 | 54.30 | 60.26 | 55.22 | 66.73 | 62.79 | 67.02 | 66.46 |
| + ChatEmb | 64.73 | 60.86 | 66.91 | 62.74 | 57.48 | 54.49 | 60.59 | 56.17 | 67.63 | 63.27 | 66.83 | 65.99 |
| + ChatEntro | 65.60 | 61.67 | 68.22 | 63.68 | 58.60 | 55.39 | 61.07 | 56.79 | 68.20 | 65.00 | 67.93 | 66.61 |
| + ZGPTDA | 66.37 | 62.04 | 68.57 | 64.02 | 59.33 | 55.77 | 61.71 | 57.62 | 68.96 | 65.32 | 68.49 | 67.54 |
| AlBert | 59.36 | 55.65 | 58.35 | 56.14 | 50.99 | 50.20 | 55.10 | 51.35 | 57.07 | 58.72 | 60.99 | 59.43 |
| + ENDA | 62.30 | 59.23 | 61.42 | 58.58 | 53.06 | 51.78 | 57.20 | 52.81 | 61.65 | 62.72 | 63.79 | 62.08 |
| + SCDA | 64.68 | 61.57 | 63.76 | 59.82 | 55.48 | 54.07 | 58.41 | 55.52 | 63.45 | 63.96 | 66.06 | 63.78 |
| + EDA | 62.88 | 60.24 | 61.90 | 58.79 | 55.29 | 53.03 | 58.06 | 54.46 | 61.99 | 63.19 | 65.47 | 63.43 |
| + Ziteg | 63.36 | 60.34 | 62.07 | 59.73 | 55.41 | 53.94 | 58.27 | 55.17 | 62.00 | 64.03 | 65.83 | 63.90 |
| + AugGPT | 64.92 | 61.61 | 63.99 | 60.82 | 56.77 | 55.54 | 61.08 | 57.07 | 63.94 | 65.31 | 67.38 | 65.24 |
| + GENCO | 64.94 | 61.69 | 64.04 | 60.88 | 56.86 | 55.58 | 61.13 | 57.07 | 63.95 | 65.35 | 67.41 | 65.30 |
| + ZGPTDA(random) | 65.19 | 62.05 | 64.07 | 61.20 | 56.71 | 55.15 | 60.81 | 56.65 | 64.02 | 65.52 | 67.42 | 67.03 |
| + ChatEmb | 64.73 | 61.82 | 64.30 | 62.14 | 55.83 | 55.17 | 59.76 | 56.79 | 63.17 | 65.26 | 66.91 | 65.98 |
| + ChatEntro | 66.12 | 63.95 | 66.06 | 63.46 | 57.77 | 56.49 | 61.28 | 58.61 | 64.65 | 66.90 | 68.64 | 67.77 |
| + ZGPTDA | 67.11 | 64.61 | 66.87 | 63.98 | 58.44 | 57.31 | 62.18 | 59.00 | 65.14 | 67.55 | 69.36 | 68.76 |
| DeBerta | 60.55 | 54.81 | 61.41 | 55.09 | 52.61 | 48.71 | 54.41 | 50.51 | 58.97 | 57.42 | 60.43 | 59.35 |
| + ENDA | 63.36 | 57.29 | 63.00 | 57.19 | 55.25 | 50.52 | 56.67 | 54.24 | 62.22 | 60.07 | 63.15 | 63.21 |
| + SCDA | 66.24 | 58.68 | 65.93 | 60.33 | 56.74 | 53.48 | 57.61 | 56.98 | 65.67 | 63.07 | 65.65 | 65.74 |
| + EDA | 64.16 | 57.96 | 65.02 | 59.20 | 56.52 | 53.02 | 57.65 | 55.71 | 63.53 | 61.40 | 63.77 | 64.23 |
| + Ziteg | 64.51 | 58.55 | 65.04 | 59.35 | 57.07 | 53.42 | 58.35 | 56.04 | 64.43 | 62.20 | 65.00 | 65.04 |
| + AugGPT | 65.73 | 60.05 | 66.04 | 60.80 | 58.90 | 55.14 | 59.41 | 57.18 | 65.50 | 63.36 | 66.82 | 66.31 |
| + GENCO | 65.83 | 60.12 | 66.11 | 60.89 | 58.95 | 55.14 | 59.49 | 57.23 | 65.51 | 63.45 | 66.86 | 66.34 |
| + ZGPTDA(random) | 65.71 | 59.63 | 66.85 | 60.80 | 58.98 | 55.57 | 59.48 | 57.37 | 65.77 | 63.48 | 66.84 | 66.79 |
| + ChatEmb | 65.13 | 60.69 | 67.97 | 60.30 | 59.05 | 55.38 | 59.36 | 56.69 | 65.63 | 64.08 | 67.55 | 67.14 |
| + ChatEntro | 66.52 | 61.66 | 69.37 | 61.54 | 59.66 | 55.85 | 60.16 | 57.60 | 67.05 | 64.44 | 68.01 | 67.95 |
| + ZGPTDA | 67.04 | 62.27 | 69.85 | 61.95 | 60.43 | 56.73 | 60.95 | 58.39 | 67.56 | 65.26 | 68.68 | 68.90 |

Table 4 confirms the undeniable strength of ZGPTDA, enhancing the performance of four models across three datasets to noteworthy levels, surpassing the original models by a significant margin of 7-10% in both Acc and F1, as seen in Possibility, especially notable in Bert and RoBerta. Furthermore, ZGPTDA demonstrates a dominant performance, significantly outperforming recent approaches such as AugGPT and GENCO. More importantly, all results lead to the following conclusions: (1) The superior performance of ZGPTDA over AugGPT highlights the varying degrees of suitability from an experimental perspective among samples generated by LLMs, indicating that not all generated samples are equally valuable. (2) ZGPTDA outperforming ZGPTDA(random) validates the necessity and effectiveness of applying scaling laws to intervene in the generated sample set, confirming that a targeted selection based on these laws enhances the utility of augmented data. (3) The advantage of ZGPTDA over ChatEmb underscores the efficacy of concatenating enhanced data with the original dataset, compared to embedding operations, suggesting that direct augmentation strategies are better. (4) ZGPTDA's superiority to ChatEntro proves the effectiveness of combining Z-number mechanisms with scaling laws, showcasing its competitiveness and utility.

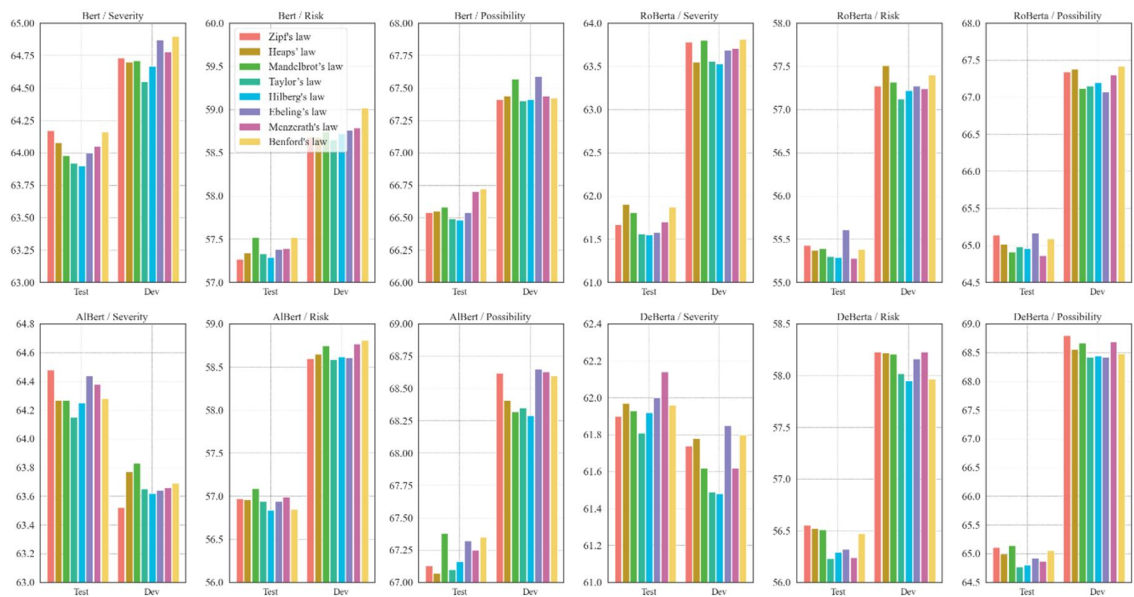

Fig.3: Ablation studies on individual law, illustrating the F1 performance of the models following the sequential exclusion of each specific law within ZGPTDA.

Additionally, Fig.3 demonstrates that the performance of the model declines if any of the laws are omitted. This observation strongly validates the effectiveness of each law individually and confirms their collective contribution to ZGPTDA. Furthermore, ZGPTDA's performance undergoes a relatively more significant decrease when either Hilberg's law or Taylor's law is abandoned. This indicates that these two laws provide more assistance to the model in achieving text classification. We hypothesize that this could be due to the high information content inherent in hazard events (as there is a need to document as many details of the hazard evolution as possible within a limited space [27], which aligns with the characterizations of information from the perspectives of entropy and density by these two laws. This finding could inspire subsequent research among peers, such as prioritizing Hilberg's law and Taylor's law in feature engineering to optimize efficiency.

Furthermore, Fig.4 illustrates that omitting any metric in ZGPTDA results in a decline in the model's classification effectiveness. This experimental outcome underscores the significance of each metric. It is noteworthy that the performance deterioration of ZGPTDA does not show a clear inclination towards any specific metric, suggesting that there is no discernible hierarchy in the effectiveness among the different metrics.

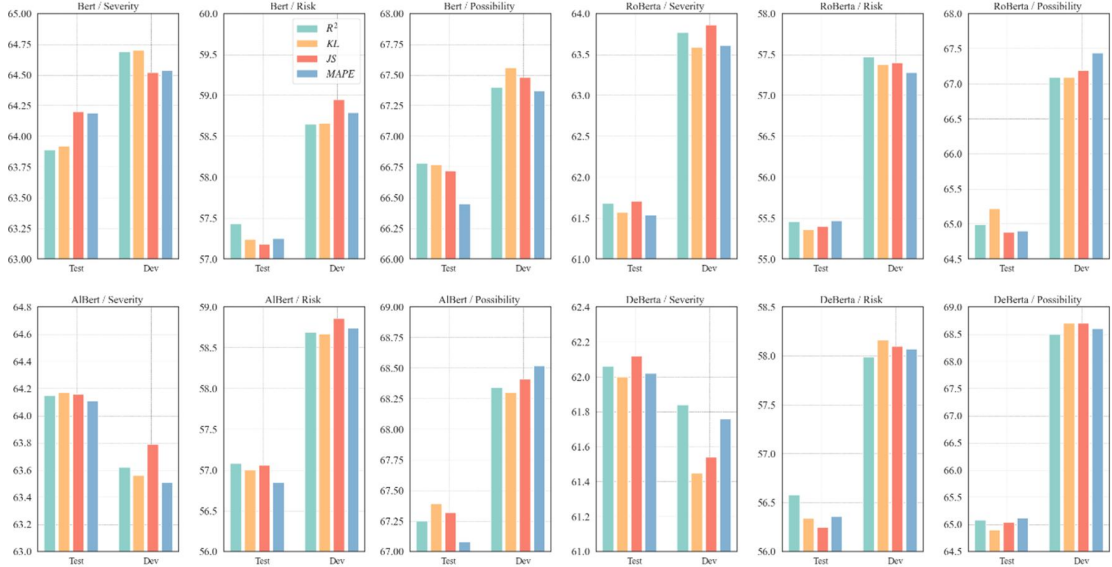

Fig.4: Ablation studies on individual metric, illustrating the F1 performance of the models following the sequential exclusion of each specific metric within ZGPTDA.

## 5 DISCUSSION

In this paper, the integration of scaling laws with natural language to empirically innovate and iterate AI methodologies for aligning LLMNL with HNL marks a novel interdisciplinary approach. This wisdom combines the rigors of quantitative research with the insights of NLP, enhancing the application of LLMs across a broader array of tasks. Additionally, this study could echo a deeper understanding of LLM training, such as how the scaling laws of parameters determine the size of LLM [45].

The discussion also ventures into why LLMs have yet to achieve the complexity (fractal) of human language. Human language is often intertwined with specific styles, biases, and deliberate errors, and is even influenced by cultural contexts [33], appearing intuitive yet not simple. These aspects are challenging for LLMs to fully replicate, as they do not possess such personal tendencies and are less prone to the kind of logical fallacies or inconsistencies humans might unintentionally introduce, given their probabilistic computations.

To illustrate this discussion concretely, let's consider the language style as an example. We launch their Type-token ratio [46], Syntactic variability (approximated through the distribution of part-of-speech tags calculated on each word) [47], and Readability, which covers the Flesch reading ease (higher scores indicate passages that are easier to read) and Dale-chall score (lower scores indicate easier readability) [48]. The results, see Table 5, notably highlight that HNL exhibit higher readability, notwithstanding the proximity of other aspects. More crucially, we explore the semantic richness, operationalized through the quantification of synonym counts per word [49], and sentiment intensity, discerned through sentence-pair sentiment analysis conducted with Bert [50], visualized in Fig.5. This unveils a richer semantic layering of HNL as opposed to LLMNL, despite a close alignment in their sentiment valences.

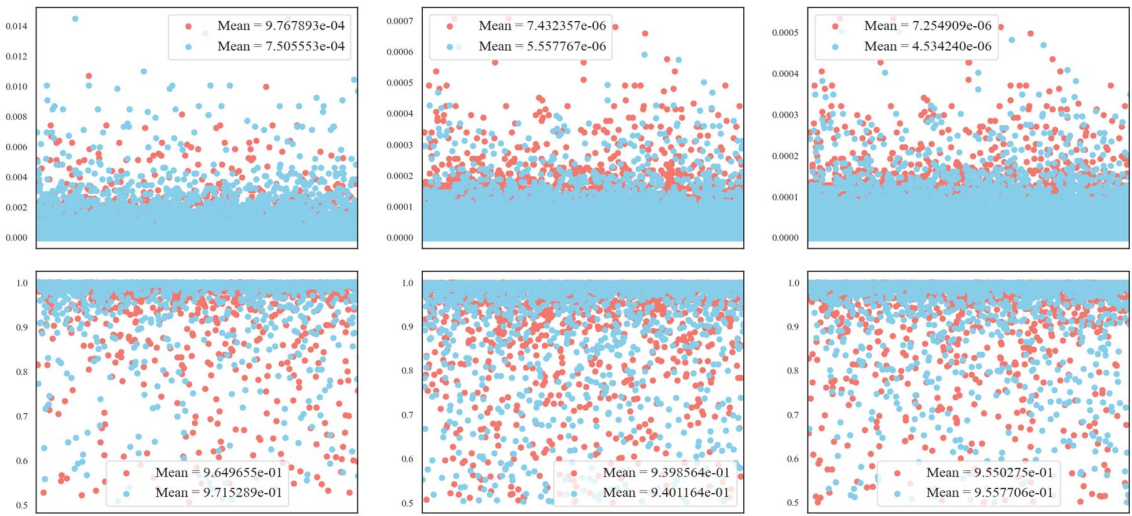

Fig.5: Results of semantic richness, sentiment intensity (row 1-2); column 1-3: Cgtd, Cheat, HC3; blue-red points: LLM, Human.

Table 5: Results of partial language styles.

| Dataset / Type | Type-token ratio | Syntactic variability | Flesch reading ease | Dale-chall score |
| --- | --- | --- | --- | --- |
| Cgtd / LLM | 0.12577 | 0.00580 | 48.54 | 5.73 |
| Cgtd / Human | 0.14884 | 0.00457 | 61.36 | 5.71 |
| Cheat / LLM | 0.02386 | 0.00033 | 24.88 | 1.16 |
| Cheat / Human | 0.04340 | 0.00038 | 32.94 | 1.31 |
| HC3 / LLM | 0.03234 | 0.00028 | 58.01 | 1.20 |
| HC3 / Human | 0.01800 | 0.00030 | 69.72 | 1.09 |

These realizations not only bolster the explanation for the lower fractal complexity of LLMNL but also potentially facilitate the advancement of human feedback-enhanced reinforcement learning (RLHF) within AI [25]. While directly embedding multifractal insights into RLHF processes is non-trivial, optimizing for more natural language feedback is highly pertinent and can guide the development of more adaptable and effective LLMs. Moreover, our methodology could be instrumental in developing AI systems that engage with the human world in a more ethically responsible manner, provided there is a deliberate shift in focus towards the moral and ethical dimensions of natural language use. For example, we can investigate the representation of unethical generated content within the parameters of scaling laws and, based on this, further develop corresponding monitoring systems, among other initiatives. Similar research has already achieved some success in the field of computer vision [51].

A limitation within ZGPTDA is the current difficulty in explicitly identifying the factors contributing to the suitability of GPT-4 generated samples, as these factors are often implicit. This challenge stems from the uncertainty in pinpointing exactly what makes a generated sample suitable or not for a given application. The difficulty lies not only in the vast range of potential variables that could influence suitability, but also in the probabilistic and somewhat opaque manner in which GPT-4 generate text.

Future research is encouraged to explore this domain further, aiming to develop methodologies or analytical tools that can assess the multitude of factors influencing the suitability of LLM-generated samples. Such advancements would greatly enhance our ability to selectively utilize generated content in a manner that maximizes relevance, effectiveness, and ethical alignment with the intended applications, thereby overcoming one of the current limitations faced by ZGPTDA and similar methodologies in the field of NLP and AI.

# 6 CONCLUSION

The advent of LLMs appears to blur the linguistic boundaries between humans and machines, yet empirically, this remains uninvestigated. To address this gap, we introduce a new methodology based on scaling laws to elucidate the differences between LLMNL and HNL, including Zipf's law, Heaps' law, Taylor's law, Hilberg's law, Ebeling's law, Menzerath's law, Benford's law, and Mandelbrot's law. Through extensive experiments across three datasets, utilizing a multitude of metrics, we have demonstrated a parity between LLMNL and HNL. Building on this foundation, we introduce ZGPTDA, an innovative data augmentation methodology tailored for few-shot text classification. ZGPTDA leverages the consistency of GPT-4 generated samples with these scaling laws, utilizing a Z-number mechanism to guide the decision-making process for data augmentation. Extensive validation in real-world scenarios has confirmed the effectiveness of ZGPTDA. We disclose a constellation of new insights. Our scrutiny meticulously unravels the subtle absence of fractal complexity inherent within LLMNL. This dissection, aided by linguistic stylistic elements such as readability, sentiment, and semantics, provides a profound augmentation of our comprehension regarding the intrinsic attributes of LLM and their prospective evolutionary trajectories. Furthermore, our empirical investigations reveal that LLM-generated datasets exhibit heterogeneous training valences, advocating for the adoption of sophisticated methodologies predicated on scaling laws and fuzzy logic to adeptly navigate these variations. Additionally, our exploration substantiates that the strategic implementation of Hilberg's law and Taylor's law can better amplify the performance of text classification. This revelation harbors the potential to galvanize scholarly pursuits, advocating for the prioritization of these laws within the ambit of feature engineering to catalyze enhanced efficiency and refine the paradigms of natural language processing. This paper sets a paradigm for more robust and reliable LLM-based NLP, paving the way for advancements that harness the potential of AI with greater confidence.